\documentclass{article}

 \usepackage{graphicx}
\usepackage{caption}
\usepackage{amsmath}
\usepackage{tikz}
\usetikzlibrary{positioning, arrows.meta, shapes, calc}
\usepackage{subcaption}
\usepackage{subcaption}
\usepackage{booktabs}
\usepackage{graphicx}
\usepackage{array} % For better table column control
\usepackage{threeparttable} % Add this in your preamble
\usepackage{tabularx}
\usepackage{wrapfig} % put this in the preamble
\usepackage{multirow} % add this in the preamble if not already there

\usepackage[utf8]{inputenc} % allow utf-8 input
\usepackage[T1]{fontenc}    % use 8-bit T1 fonts
\usepackage{hyperref}       % hyperlinks
\usepackage{url}            % simple URL typesetting
\usepackage{booktabs}       % professional-quality tables
\usepackage{amsfonts}       % blackboard math symbols
\usepackage{nicefrac}       % compact symbols for 1/2, etc.
\usepackage{microtype}      % microtypography
\usepackage{xcolor}         % colors

\title{WindMiL: Equivariant Graph Learning for Wind Loading Prediction}

\usepackage[sglblindworkshop, final]{neurips_2025}
\workshoptitle{New Perspectives in Graph Machine Learning}
\author{
  Themistoklis Vargiemezis$^{1}$ \\
  \texttt{tvarg@stanford.edu}
  \And
  Charilaos Kanatsoulis$^{2}$ \\
  \texttt{charilaos@cs.stanford.edu}
  \And
  Catherine Gorl{\'e}$^{1}$ \\
  \texttt{gorle@stanford.edu}
  \and \\
  $^{1}$Department of Civil \& Environmental Engineering, Stanford, CA, USA \\
  $^{2}$ Department of Computer Science,  Stanford, CA, USA \\
}

\begin{document}

\maketitle

\begin{abstract}
 Accurate prediction of wind loading on buildings is crucial for structural safety and sustainable design, yet conventional approaches such as wind tunnel testing and large-eddy simulation (LES) are prohibitively expensive for large-scale exploration. Each LES case typically requires at least 24 hours of computation, making comprehensive parametric studies infeasible. We introduce \textsc{WindMiL}, a new machine learning framework that combines systematic dataset generation with symmetry-aware graph neural networks (GNNs). First, we introduce a large-scale dataset of wind loads on low-rise buildings by applying signed distance function interpolation to roof geometries and simulating 462 cases with LES across varying shapes and wind directions. Second, we develop a reflection-equivariant GNN that guarantees physically consistent predictions under mirrored geometries. Across interpolation and extrapolation evaluations, \textsc{WindMiL} achieves high accuracy for both the mean and the standard deviation of surface pressure coefficients (e.g., RMSE $\leq 0.02$ for mean $C_p$) and remains accurate under reflected-test evaluation, maintaining hit rates above $96\%$ where the non-equivariant baseline model drops by more than $10\%$. By pairing a systematic dataset with an equivariant surrogate, \textsc{WindMiL} enables efficient, scalable, and accurate predictions of wind loads on buildings.
\end{abstract}

\section{Introduction}
% \CK{and set the ground for machine learning research in the wind engineering field? or the particular problem. In some sense we want to emphasize that we set up the ML paradigm for this particular problem}

Wind loading on buildings is crucial for structural engineering and resilient urban planning. Accurate predictions of surface pressure loads and integrated loads guide both safety codes and sustainable building practices. Yet, traditional approaches such as wind tunnel testing and full-scale experiments are costly and time-consuming \citep{alrawashdeh2015wind,richards2012pressures,vargiemezis2023experimental, vargiemezis2024modeling}, while high-fidelity CFD simulations, such as large-eddy simulation (LES), though accurate, remain prohibitively expensive for large-scale design exploration \citep{potsis2023computational,blocken2015computational,vargiemezis2025predictive,vargiemezis2024urban}. These limitations highlight the need for efficient and accurate surrogates that can operate at scale.

Recent advances in geometric deep learning provide a powerful foundation for such surrogates. Graph neural networks (GNNs) naturally represent the irregular meshes common in CFD, with nodes as mesh points and edges as connectivity. Models such as MeshGraphNets \citep{pfaff2020learning} and Graph Network Simulators \citep{sanchez2020learning} show the potential of graph-based learning for fluid and structural dynamics, while PolyGNN \citep{chen2024polygnn} highlights their use in building geometry reconstruction. These works highlight that graph learning can bridge building geometry, flow physics, and predictive modeling.

\begin{wrapfigure}{r}{0.32\textwidth}
    \centering
    \vspace{-5mm} % adjust vertical space above
    \includegraphics[width=0.32\textwidth]{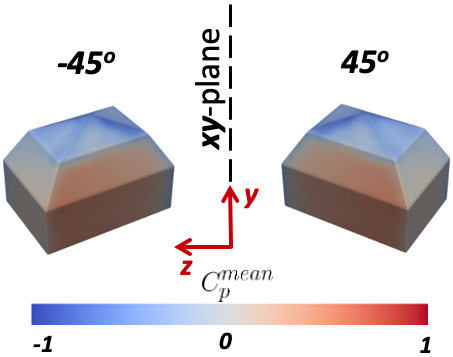}
    \vspace{-5mm} % adjust space between image and caption
    \caption{Contour plots of mean $C_p$ at $-45^{\circ}$ and $-45^{\circ}$ wind incidence.}
    \vspace{-5mm} % adjust vertical space above
    \label{fig:cp_symmetry}
\end{wrapfigure}

To fully realize the potential of graph learning in wind engineering, surrogate models must incorporate the physical symmetries that govern flows around buildings. Reflectional and rotational invariances are fundamental for ensuring consistent and reliable predictions. For example, when the flow is aligned with $x-$direction, $y$ is the vertical, and $z$ is the spanwise, the pressure distribution of a building at $+45^{\circ}$ wind incidence should be the reflected version of a building at $-45^{\circ}$ wind incidence with respect to the $xy-$plane, as shown in Fig. \ref{fig:cp_symmetry}.

Equivariant graph learning embeds these symmetry constraints directly into the model, ensuring that its predictions transform consistently under reflections and rotations. Approaches such as group-equivariant CNNs \citep{cohen2016group} and E(3)-equivariant GNNs like NequIP \citep{batzner20223} have already demonstrated substantial gains in efficiency and generalization in physical sciences. These advantages make equivariant GNNs a suitable modeling choice for wind loading surrogates.

Progress in this direction has been hindered by the lack of systematic datasets. While benchmark experiments such as the TPU database \citep{tpu2007} remain invaluable, they cover only a small number of canonical building geometries. By contrast, other areas such as automotive aerodynamics have advanced through parametric dataset generation paired with CFD \citep{benjamin2025systematic}. No comparable dataset exists for wind loading on low-rise buildings, despite their importance to building safety.

\textbf{Our contribution}: We introduce \textsc{WindMiL}, a new machine learning paradigm for wind loading prediction that combines systematic datasets with symmetry-aware graph learning. Specifically:

\begin{enumerate}
    \item \textbf{Dataset}. We generate a large-scale dataset of wind loading on low-rise buildings, with systematically varying roof morphologies and wind directions, using LES. This extends databases such as the TPU database through controlled geometric interpolation.
    \item \textbf{Model}. We develop a reflection-equivariant GNN that respects reflectional invariances, providing physically consistent surrogates. Our approach achieves error reduction by more than 10\% on the symmetrical geometries compared to the non-equivariant baseline.
\end{enumerate}

By releasing both the dataset and the model, we establish an ML paradigm for wind engineering that is data-driven, symmetry-aware, and scalable. This paradigm opens new opportunities for accurate, efficient, and physically consistent prediction of wind loads, advancing resilient and sustainable urban design.

\section{Dataset generation} 
% Geometry parameterization:
% Define low-rise building archetypes (rectangular, L-shaped, pitched roofs).
% Systematic morphing via signed distance functions (SDF) interpolation (adapted from Benjamin & Iaccarino).

% Flow simulations:
% Large-eddy simulations (LES) for high-fidelity results
% Wind directions: multiple yaw angles (e.g., 0 to +90°). with data augmentation/symmetry 
% Quantities of Interest (QoIs): surface pressure coefficients, integrated loads.
% Dataset statistics: number of buildings, wind angles, total cases.
% (Optional Figure): Barycentric interpolation map + examples of generated shapes.
\subsection{Geometry parameterization}
\begin{wrapfigure}{r}{0.4\textwidth}
    \centering
    \vspace{-7mm} % adjust vertical space above
    \includegraphics[width=0.4\textwidth]{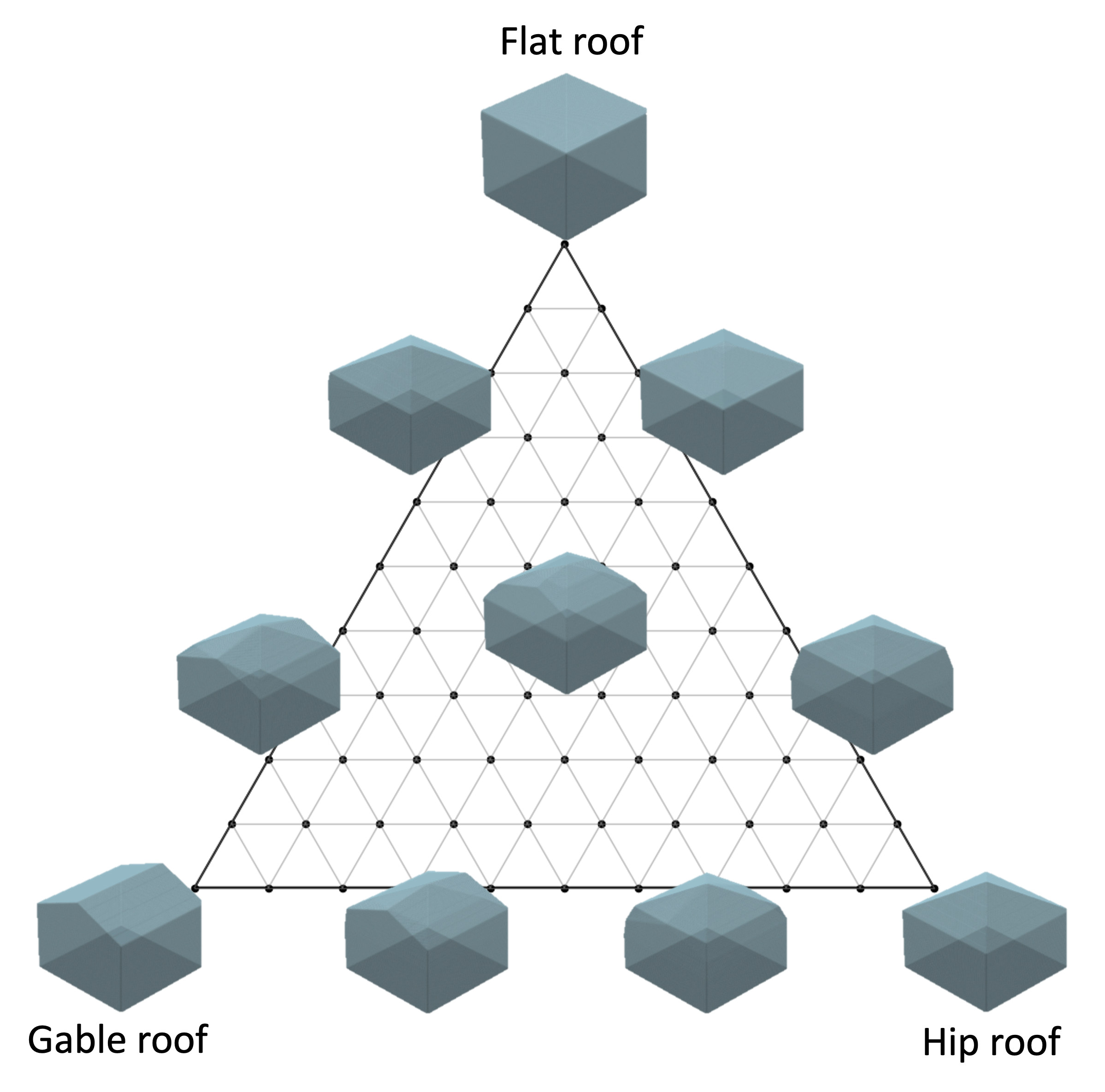}
    \vspace{-3mm} % adjust space between image and caption
    \caption{Convex hull of the three basis geometries. Points inside the convex hull correspond to interpolated buildings.}
    \label{fig:barycentric_map}
    \vspace{-10mm} % adjust vertical space above
\end{wrapfigure}

In this work, we adapt the Signed Distance Function (SDF)-based interpolation method, originally developed for automotive geometries by \cite{benjamin2025systematic}, to systematically generate diverse building shapes for wind loading analysis. We begin with three basis geometries derived from the TPU dataset \citep{tpu2007}: a flat roof, a gable roof, and a hip roof. All buildings share the same footprint of 18 m length × 12 m width in full scale, while their heights differ: 8 m for the flat roof, 16 m for the gable roof, and 24 m for the hip roof.

Each geometry is first converted into a binary grid using ray-tracing, where cells are marked as occupied or empty based on ray–surface intersections. The binary grids are then transformed into signed distance functions (SDFs), which assign to each grid point the signed distance to the building surface. By convention, the SDF equals zero on the surface, is positive outside, and negative inside. This continuous representation enables smooth and consistent interpolation of shapes. To generate new buildings, we construct a convex hull in the SDF space spanned by the three basis geometries, as shown in Fig. \ref{fig:barycentric_map}. Barycentric interpolation within this convex hull produces intermediate SDFs that smoothly transition between the flat, gable, and hip roof cases. For the wind incidence at $0^\circ$, this yields 66 unique interpolated buildings. To build a complete database, each interpolated building is rotated in $15^\circ$ increments from $0^\circ$ to $90^\circ$, resulting in $66 \times 7 = 462$ cases, with one convex hull generated for each wind direction.

\begin{figure}[ht!]
\centering
\includegraphics[width=0.9\textwidth]{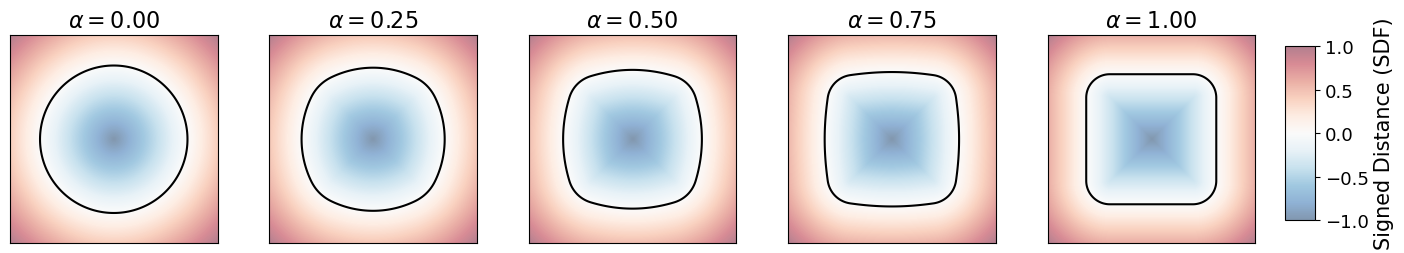}
\vspace{-2mm}
\caption{SDF interpolation between a sphere and a cube, per Eq. (1) for different $\alpha$.}
\label{fig:sdf_workflow}
\end{figure}

Formally, interpolation between two SDFs $\phi_1(\vec{x_i})$ and $\phi_2(\vec{x_i})$ on a structured grid, where $(\vec{x_i})$ are spatial coordinates on a discrete grid, can be written as
\begin{equation}
\phi_3((\vec{x_i})) = \alpha \phi_1((\vec{x_i})) + (1 - \alpha) \phi_2((\vec{x_i})),
\end{equation}
where $\alpha$ is a scalar weight. Varying $\alpha$ yields intermediate geometries between the basis cases.  Applying Eq. (1) to a unit sphere and unit cube, one can produce intermediate shapes by varying $\alpha$, an example of which is shown in Fig. \ref{fig:sdf_workflow}

Finally, the interpolated SDFs are reconstructed into surface meshes using the marching cubes algorithm \citep{benjamin2025systematic}, followed by Laplacian smoothing to refine geometry quality. This ensures the resulting buildings are physically suitable for downstream tasks such as CFD simulations or data-driven surrogate modeling. By combining convex-hull interpolation with systematic rotations, the method enables the creation of a large dataset of building geometries.

\subsection{Large-eddy Simulations}
To generate the dataset, LES simulations are performed using the CharLES code, a low-Mach, isentropic solver developed by \cite{charles}. CharLES implements a finite volume approach with an automated body-fitted meshing technique based on 3D-clipped Voronoi diagrams, and uses numerical schemes with second-order accuracy in space and time to solve the governing equations \citep{ambo2020aerodynamic}. In addition, the Vreman turbulence model is used to model the unresolved part of the stress tensor \citep{vreman2004eddy}. CharLES has been extensively validated against experimental data in a range of wind engineering applications. For instance, it has been validated against wind tunnel measurements of wind-induced pressure loads on high- and low-rise buildings \citep{ciarlatani2023investigation,vargiemezis2024urban, vargiemezis2025predictive}, as well as against measurements of wind-driven natural ventilation~\citep{hwang2022large}. Comparisons against field measurements have also been performed for wind pressures on the Space Needle \citep{HOCHSCHILD2024105749} and for natural ventilation flow in a dense urban environment~\citep{hwang2023large}. In addition, CharLES has been used to create datasets for predicting wind fields in urban areas using a U-net-based model \citep{vargiemezis2025large}.

\begin{figure}[htpb]
    \centering
    \includegraphics[width=1\columnwidth]{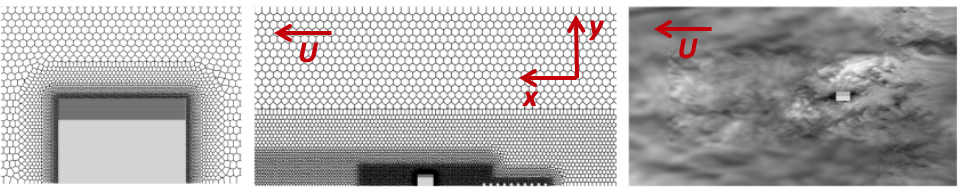}
    \caption{View of the mesh used for the large-eddy simulations, with side views of refinement zones shown: (left) near the building and (middle) in the surrounding area. (right) Top view of instantaneous velocity magnitude contours showing the turbulent flow around the building.}
    \vspace{-2mm}
    \label{fig:domain_flow_pred}
    \vspace{-2mm}
    \end{figure}

The mesh uses $2.1 \times 10^6 $ control volumes, with refinement near the no-slip surfaces of the building, as shown in Fig. \ref{fig:domain_flow_pred}. The resolution on the building surfaces is $\Delta / H_{ref} = 0.0098$, where $\Delta$ is the mesh size locally, and $H_{ref}$ is the characteristic height of the building. The domain extents $5H_{ref}$ upstream, $15H_{ref}$ downstream, and $5H_{ref}$ in the lateral and vertical directions, following the proposed guidelines \citep{franke2011cost}. At the inlet, a logarithmic mean velocity profile is prescribed, and artificially generated turbulent fluctuations are superimposed. The turbulent velocity field is generated using the divergence-free digital filter method proposed by \citep{kim2013divergence}. At the outlet, a zero gradient condition is applied, while the two lateral boundaries are periodic, and a slip condition is applied at the top boundary. Finally, at the ground, a rough wall function for a neutral atmospheric boundary layer with a fixed roughness of 0.0027 m is specified. More details regarding the setup can be found in the study of a different isolated building, since the same setup is used ~\citep{vargiemezis2025predictive}.

The simulation is run for a total time of 20 seconds, and statistics are collected for the last 15 seconds, after the initial burn-in period of 5 seconds. The total time for one simulation on 64 CPUs requires 24 hours. The quantities of interest (QoIs) of the simulation are the time-averaged (mean) and the standard deviation (std) of the pressure coefficient. The pressure coefficient at a surface node $i$ is defined as
\begin{equation}
    C_{p,i} = \frac{p_i - p_{\infty}}{\tfrac{1}{2}\rho U_{\infty}^2},
\end{equation}

where $p_i$ is the local surface pressure at location $i$ on the building surface ,$p_{\infty}$ is the freestream pressure, $\rho$ is the density of air, and $U_{\infty}$ is the reference velocity.

\section{Model architecture}
% Graph representation:
% Mesh nodes = building surface panels.
% Edges = adjacency matrix.
% Node features = surface normals, wind angle features.

% Symmetry equivariance:
% Group = reflection symmetry across YZ-plane (mirror).
% Model enforces equivariance in pressure/load predictions.
% Baseline: standard GNN (GCN/GraphSAGE).
% Our model: Symmetry-equivariant message passing + invariant aggregation?????
% (Figure): schematic of symmetry-equivariant GNN.

We represent each building surface as a point cloud, from which we construct a graph $\mathcal{G} = (\mathcal{V}, \mathcal{E})$, where nodes $v \in \mathcal{V}$ correspond to sampled surface points and edges $e \in \mathcal{E}$ are defined based on spatial proximity. Each node is associated with a set of geometric features $\mathbf{X} \in \mathbb{R}^{|\mathcal{V}| \times d}$. In particular, each node $v$ has features $\mathbf{x}_v \in \mathbb{R}^d$, with $d=6$: normalized coordinates $(x,y,z)$ with respect to building height $H_{ref}$ and surface normal components $(n_x,n_y,n_z)$, which are unit vectors oriented outward of building surface. The learning task is to predict the per-node mean or std pressure coefficients $C_p$.
To process the graph structure and the geometrical features of each building, we employ a message-passing GNN $f_\theta\left(\mathcal{G},\mathbf{X}\right)$, \citep{kipf2016semi, xu2018powerful,hamilton2017inductive}, defined by the following recursive formula:
\begin{align}\label{eq:GNNrec00}
   \mathbf{x}_v^{(l)} = h^{(l-1)}\left(\mathbf{x}_v^{(l-1)},g^{(l-1)}\left(\left\{\mathbf{x}_u^{(l-1)}:u\in\mathcal{N}\left(v\right)\right\}\right)\right),
\end{align}
where $\mathcal{N}(v)$ represents the 1-hop neighborhood of vertex $v$, and $g,h$ are permutation equivariant operators. In our implementation, $f_\theta\left(\mathcal{G},\mathbf{X}\right)$ is modeled as a stack of GraphSAGE layers \citep{hamilton2017inductive} with residual connections, layer normalization, and a global skip from the input projection.

Wind loading exhibits a reflection symmetry with respect to the horizontal ($xy$) plane, as shown in Fig. \ref{fig:cp_symmetry}. For instance, the surface pressure distribution at a wind incidence angle of $+45^\circ$ is the reflected version of that at $-45^\circ$, when the reflected building is taken across the $xy$ plane.  To encode this invariance, we define a reflection operator $\mathcal{R}$ that flips the vertical coordinate and normal component:
\[
\mathcal{R}:\ (z,n_z) \mapsto (-z,-n_z),
\]
while leaving $(x,y,n_x,n_y)$ unchanged. The model is designed such that predictions are invariant under this transformation.  
To enforce symmetry, both the original and reflected features are passed through the same encoder $f_\theta$. The resulting embeddings are then averaged:
\[
    \mathbf{z}_v = \tfrac{1}{2}\Big(f_\theta(\mathcal{G},\mathbf{X})[v] + f_\theta(\mathcal{G}, \mathcal{R}(\mathbf{X}))[v]\Big).
\]

The overall model $F(\cdot)$, which maps inputs to predictions, then satisfies the reflection-equivariant property $F(\mathcal{R}(\mathbf{x})) = F(\mathbf{x})$.
Finally, embeddings $\mathbf{z}_v$ are passed through a feed-forward predictor $g_\phi$, yielding the node-level outputs $C_{p,v}=g_\phi(\mathbf{z}_v)$. This architecture ensures predictions that respect the underlying physical symmetry of wind loading on buildings. 

\section{Experiments}
\begin{figure}[ht]
    \centering
    \begin{subfigure}[c]{0.45\textwidth}
        \centering
        \includegraphics[width=\linewidth]{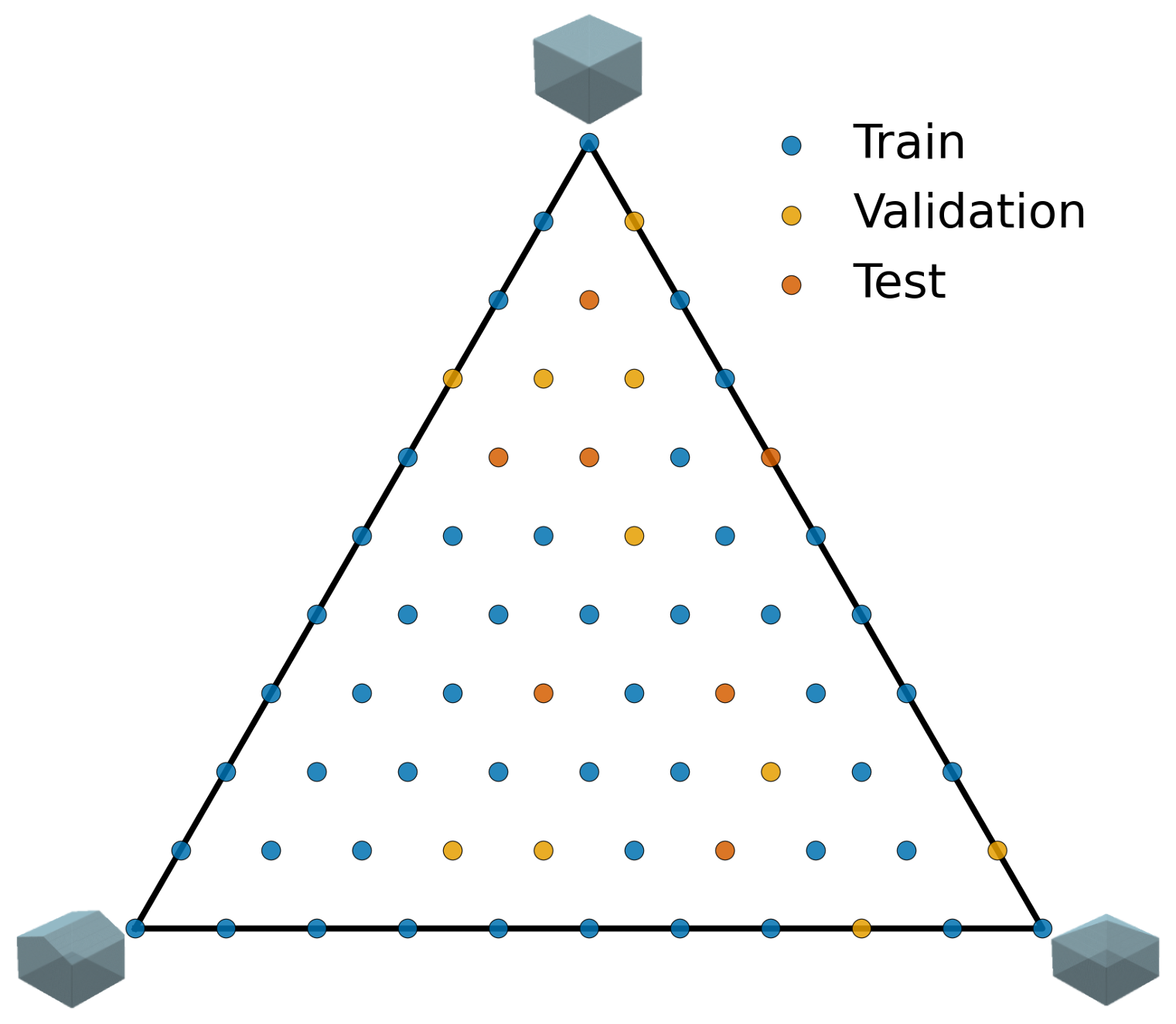}
        \caption{Random interpolation split. Train/dev/test are assigned randomly.}
        \label{fig:random-visualization}
    \end{subfigure}
    \hfill
    \begin{subfigure}[c]{0.45\textwidth}
        \centering
        \includegraphics[width=\linewidth]{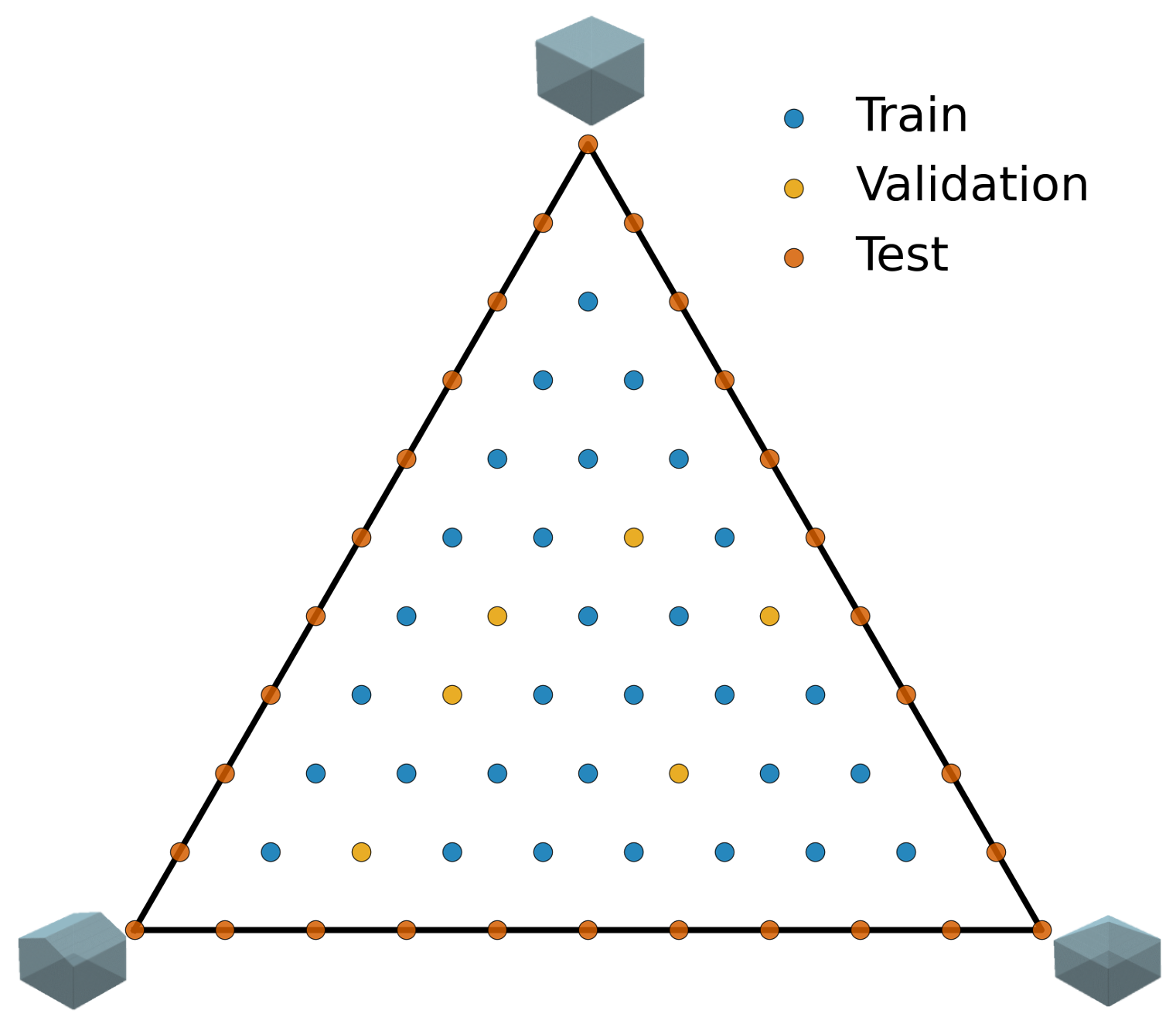}
        \caption{Extrapolation split. Geometries on the convex hull boundary are held out for testing.}
        \label{fig:extrap-visualization}
    \end{subfigure}
    \vspace{-1mm}
    \caption{Visualization of dataset splits in the barycentric shape space. 
    The convex hull is spanned by the three basis roof geometries (flat, gable, hip), and each point corresponds to one interpolated building.}
    \label{fig:dataset-splits}
    \vspace{-3mm}
\end{figure}

We compare the reflection-equivariant GNN, \textsc{WindMiL}, with the baseline model \textsc{GraphSage}, which does not incorporate geometric symmetries. Both models are evaluated in two cases: (a) interpolation and (b) extrapolation in the shape space, see also Fig.~\ref{fig:dataset-splits}.  Interpolation corresponds to a random train/dev/test split of the points in the convex hull, while extrapolation is assessed on unseen building configurations. In extrapolation, geometries located on the convex hull boundary of the shape space are left out for testing, while the remaining points are randomly divided into train/dev. 

For both split methods, we report results with and without including reflected data in the test set. It is important to note that training is always performed using only the original (non-reflected) geometries. In the “+Sym” evaluation setting (see also Table \ref{tab:interp}), we extend the test set to include the reflections of the held-out test geometries. This setup evaluates whether the models generalize correctly to symmetrical geometries, without having seen them during training. We evaluate the predictive accuracy using five metrics: root mean square error (RMSE), mean squared error (MSE), mean absolute error (MAE), coefficient of determination ($R^2$), and hitrate (\%).  The hitrate is defined as the percentage of predictions within a tolerance of $\pm 0.10$ for mean $C_p$ and $\pm 0.05$ for std $C_p$, which correspond to the $10\%$ of their maximum values observed on the buildings.

\subsection{Interpolation performance}
Table~\ref{tab:interp} summarizes the interpolation results. When evaluated on the non-reflected test set, both \textsc{GraphSage} and \textsc{WindMiL} achieve high predictive accuracy for the mean and std $C_p$. Errors remain small, with RMSEs below $0.02$ for mean $C_p$ and $0.003$ for std $C_p$, and hit rates exceeding $95\%$ and $97\%$, respectively. This confirms that both models successfully capture the pressure variations within the interpolation regime.

\begin{table}[htb!]
\centering
\setlength{\tabcolsep}{5pt}
\renewcommand{\arraystretch}{1}
\footnotesize
\begin{tabular}{clccccc}
\toprule
& Method & RMSE & MSE & MAE & $R^2$ & Hitrate (\%) \\
\midrule
\multirow{4}{*}{\rotatebox{90}{Mean $C_p$}} 
 & \textsc{GraphSage} & 0.019 & 0.211 & 0.032 & 0.980 & 95.95 \\
 & \textsc{GraphSage} + Sym & 0.055 & 0.273 & 0.053 & 0.943 & 85.41 \\
 & \textsc{WindMiL} & 0.020 & 0.213 & 0.033 & 0.979 & 95.80 \\
 & \textsc{WindMiL} + Sym & 0.019 & 0.207 & 0.031 & 0.981 & 96.53 \\
\midrule
\multirow{4}{*}{\rotatebox{90}{Std $C_p$}} 
 & \textsc{GraphSage} & 0.003 & 0.130 & 0.012 & 0.958 & 97.98 \\
 & \textsc{GraphSage} + Sym & 0.008 & 0.169 & 0.020 & 0.882 & 91.63 \\
 & \textsc{WindMiL} & 0.003 & 0.137 & 0.013 & 0.949 & 97.62 \\
 & \textsc{WindMiL} + Sym & 0.003 & 0.135 & 0.013 & 0.953 & 97.86 \\
\bottomrule
\end{tabular}
\vspace{+1mm}
\caption{Interpolation performance on mean and std $C_p$. \emph{+Sym} indicates that the test set additionally includes the reflections of the test set geometries.}
\label{tab:interp}
\vspace{-4mm}
\end{table}
When the test set is extended with reflected geometries (\emph{+Sym}), however, clear differences are observed. The baseline \textsc{GraphSage} shows a substantial drop in performance, with the hitrate decreasing from $95.95\%$ to $85.41\%$ for mean $C_p$ and from $97.98\%$ to $91.63\%$ for std $C_p$. This indicates that the non-equivariant model does not generalize consistently to symmetry-transformed cases and effectively treats reflections as unseen geometries. In contrast, \textsc{WindMiL} maintains high accuracy under reflection, achieving $96.53\%$ and $97.86\%$ hitrates for mean and std $C_p$, respectively, with similar RMSE and $R^2$ values to the original test set. These results show that explicitly encoding reflection symmetry not only enforces physically consistent predictions but also improves robustness to unseen symmetric transformations.

\subsection{Extrapolation performance}
Table~\ref{tab:extra} reports results for extrapolation, where building geometries on the convex hull boundary of the shape space are held out for testing. As expected, errors increase compared to interpolation since the models are evaluated on unseen geometries outside the training distribution. For the non-reflected test set, RMSE increases to $\sim0.05$ for mean $C_p$ and $\sim0.019$ for std $C_p$, yet both \textsc{GraphSage} and \textsc{WindMiL} maintain high predictive accuracy with $R^2$ values between $0.94$–$0.98$ and hitrates above $94\%$ and $97\%$ for mean and std $C_p$, respectively. These results indicate that both models preserve their accuracy in the more challenging extrapolation task.

\begin{table}[htb!]
\centering
\setlength{\tabcolsep}{5pt}
\renewcommand{\arraystretch}{1}
\footnotesize
\begin{tabular}{clccccc}
\toprule
& Method & RMSE & MSE & MAE & $R^2$ & Hitrate (\%) \\
\midrule
\multirow{4}{*}{\rotatebox{90}{Mean $C_p$}} 
 & \textsc{GraphSage} & 0.047 & 0.227 & 0.033 & 0.976 & 94.98 \\
 & \textsc{GraphSage} + Sym & 0.076 & 0.581 & 0.054 & 0.939 & 84.67 \\
 & \textsc{WindMiL} & 0.050 & 0.251 & 0.035 & 0.974 & 94.16 \\
 & \textsc{WindMiL} + Sym & 0.049 & 0.242 & 0.034 & 0.974 & 94.69 \\
\midrule
\multirow{4}{*}{\rotatebox{90}{Std $C_p$}} 
 & \textsc{GraphSage} & 0.019 & 0.0365 & 0.012 & 0.944 & 97.35 \\
 & \textsc{GraphSage} + Sym & 0.029 & 0.086 & 0.020 & 0.870 & 91.57 \\
 & \textsc{WindMiL} & 0.019 & 0.037 & 0.013 & 0.942 & 97.30 \\
 & \textsc{WindMiL} + Sym & 0.019 & 0.037 & 0.012 & 0.944 & 97.33 \\
\bottomrule
\end{tabular}
\vspace{+1mm}
\caption{Extrapolation performance on mean and std $C_p$. \emph{+Sym} indicates that the test set additionally includes the reflections of the test set geometries.}
\label{tab:extra}
\vspace{-2mm}
\end{table}
When the test set is extended with reflected geometries (\emph{+Sym}), differences between the models are observed. The baseline \textsc{GraphSage} shows a clear decrease in performance, with hitrates dropping from $94.98\%$ to $84.67\%$ for mean $C_p$ and from $97.35\%$ to $91.57\%$ for std $C_p$, along with corresponding reductions in $R^2$. In contrast, \textsc{WindMiL} remains accurate under symmetry transformations, maintaining hit rates of $94.69\%$ for mean $C_p$ and $97.33\%$ for std $C_p$, with nearly unchanged RMSE values. This consistency highlights the benefit of explicit reflection equivariance; the model generalizes more accurately to unseen symmetric configurations, avoiding the errors observed in the non-equivariant baseline.

\subsection{Qualitative Analysis}

\begin{figure}[htb!]
    \centering
    \includegraphics[width=\linewidth]{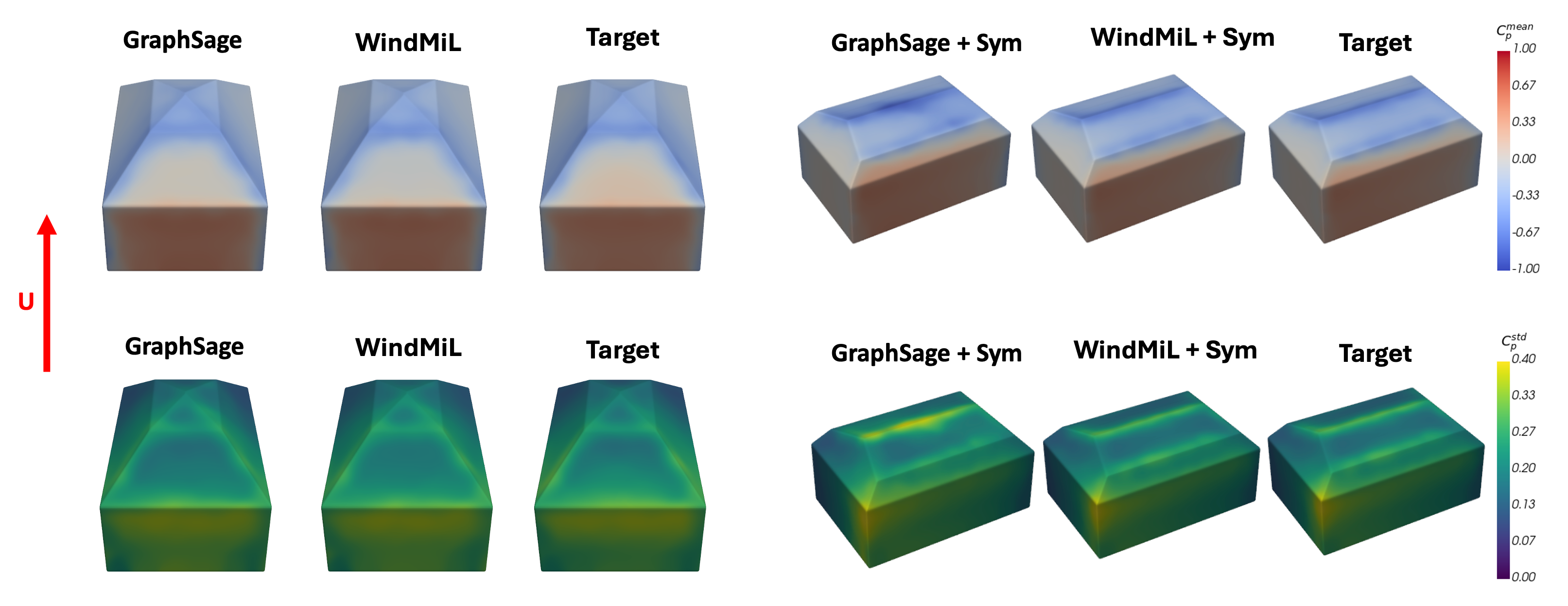}
    \caption{Contour plots of mean and std $C_p$ for the interpolation split. 
    Columns show \textsc{GraphSAGE}, \textsc{WindMiL}, and LES ground truth; rows correspond to mean $C_p$ and std $C_p$.}
    \vspace{-2mm}
    \label{fig:contours-random}
    \vspace{-2mm}
\end{figure}

\begin{figure}[htb!]
    \centering
    \includegraphics[width=\linewidth]{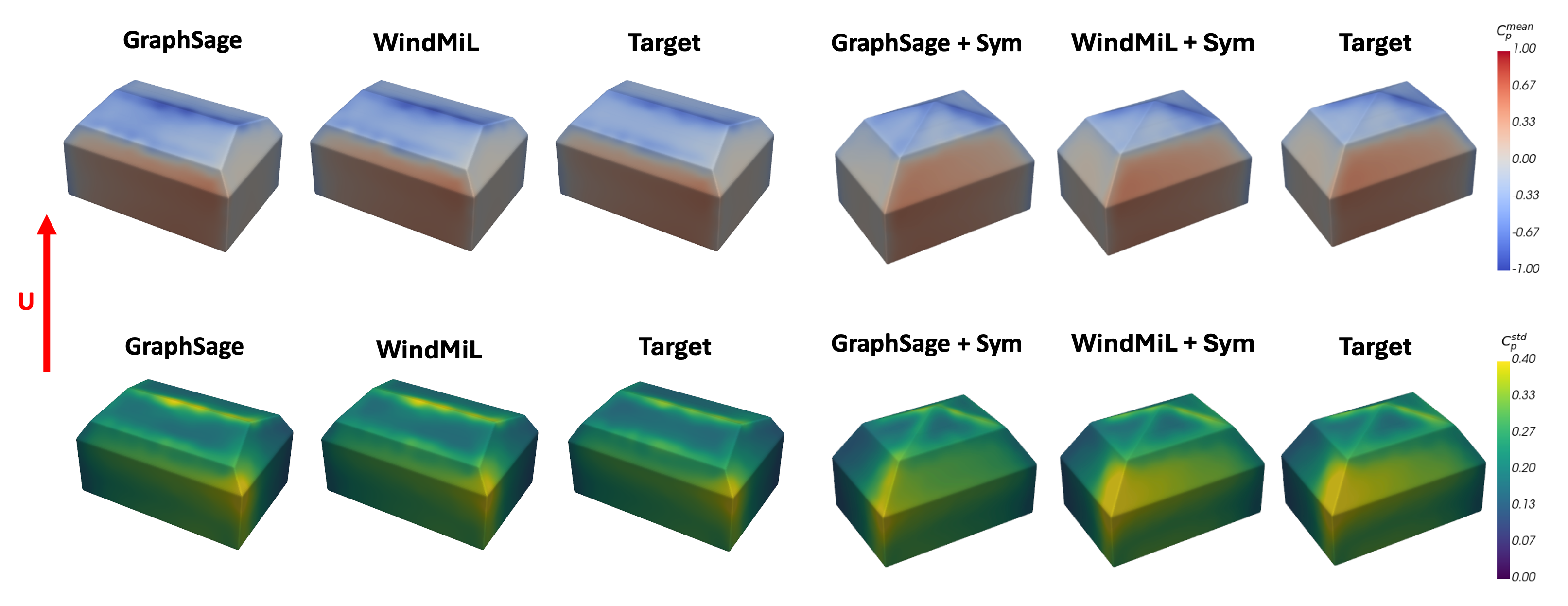}
    \caption{Contour plots of mean and std $C_p$ for the interpolation split. 
    Columns show \textsc{GraphSAGE}, \textsc{WindMiL}, and LES ground truth; rows correspond to mean $C_p$ and std $C_p$.}
    \vspace{-2mm}
    \label{fig:contours-extrap}
    \vspace{-2mm}
\end{figure}

To further assess predictive performance, we present contour plots of mean and std $C_p$ on the building surfaces for the interpolation and extrapolation cases in Fig.~\ref{fig:contours-random} and Fig.~\ref{fig:contours-extrap}, respectively. Predictions from the baseline \textsc{GraphSAGE} and our proposed \textsc{WindMiL} are compared against the LES targets.

For the standard interpolation test set in Fig. \ref{fig:contours-random}, both models capture the main spatial patterns of $C_p$. They correctly reproduce high- and low-pressure regions across roof and wall surfaces. The close visual agreement with LES is consistent with the quantitative metrics in Table~\ref{tab:interp}, where both models achieve low errors and hitrates above $95\%$. When the reflected geometries \emph{+Sym} are included, differences become more visible; the \textsc{GraphSAGE} tends to overpredict both mean and std $C_p$ on roof surfaces, while \textsc{WindMiL} maintains close agreement with the LES ground truth.

For the extrapolation split in Fig. \ref{fig:contours-extrap}, both models remain accurate for the mean $C_p$ and std $C_p$, which is confirmed with the quantitative analysis of Table \ref{tab:extra}, where both models achieved hitrates of more than $94\%$. When the $+{\rm Sym}$ extrapolation is considered, both models remain accurate for mean $C_p$, but differences are visible for std $C_p$. The \textsc{GraphSAGE} significantly underpredicts the variability at the building corners, specifically where the ground truth indicates high std values. \textsc{WindMiL}, by comparison, captures these regions much more closely. These qualitative observations show that that reflection symmetry is particularly important for predicting higher-order statistics of pressure loads.

\section{Conclusions}
We introduced \textsc{WindMiL}, a symmetry-aware graph learning framework for predicting wind loading on low-rise buildings, together with a systematically generated LES dataset of 462 buildings with varying roof types and wind directions. The dataset is built by interpolating between canonical roof geometries using signed distance functions and running high-fidelity LES for each configuration. A single LES simulation requires on the order of 24 hours on 64 CPU cores. On top of this dataset, we proposed a reflection-equivariant GNN surrogate that encodes the physical symmetry, such that pressures at $+45^\circ$ wind incidence should mirror those at $-45^\circ$. \textsc{WindMiL} achieves high accuracy for both mean and standard deviation of the surface pressure coefficient $C_p$ in both interpolation and extrapolation settings with hitrates above $95\%$. \textsc{WindMiL} achieves similar accuracy when the reflected (\emph{+Sym}) geometries are included in the test set, while the standard \textsc{GraphSAGE} degrades by more than 10 percentage points in hitrate. These results show that explicitly enforcing symmetry in the model architecture improves physical consistency without sacrificing accuracy, and enables reliable surrogate predictions at a fraction of LES cost. Future work will extend this framework to multi-building urban areas, additional inflow conditions, and the prediction of integrated loads for structural design.

%%%%%%%%%%%%
%%%%%%%%%%%%

\bibliographystyle{plainnat}
\bibliography{main.bib}

\end{document}